# Folk-ontological stances toward robots and psychological human likeness

**Edoardo Datteri**

RobotiCSS Lab, University of Milano-Bicocca, edoardo.datteri@unimib.it

https://en.unimib.it/edoardo-datteri

https://scholar.google.it/citations?user=wGwyuFkAAAAJ&hl=it

https://orcid.org/0000-0003-0323-2985

**Abstract**. It has often been argued that people can attribute mental states to robots without making any ontological commitments to the reality of those states. But what does it mean to 'attribute' a mental state to a robot, and what is an 'ontological commitment'? It will be argued that, on a plausible interpretation of these two notions, it is not clear how mental state attribution can occur without any ontological commitment. Taking inspiration from the philosophical debate on scientific realism, a provisional taxonomy of folk-ontological stances towards robots will also be identified, corresponding to different ways of understanding robotic minds. They include realism, non-realism, eliminativism, reductionism, fictionalism and agnosticism. Instrumentalism will also be discussed and presented as a folk-epistemological stance. In the last part of the article it will be argued that people's folk-ontological stances towards robots and humans can influence their perception of the human-likeness of robots. The analysis carried out here can be seen as encouraging a "folk-ontological turn" in human-robot interaction research, aimed at explicitly determining what beliefs people have about the reality of robot minds.

## 1. Introduction

It has long been observed that people occasionally attribute mental states, including beliefs, desires, and intentions, to robots. It has also been claimed that mental state attribution does not imply believing in the reality of these mental states: one may attribute, say, a belief to a robot, without believing that the robot really has beliefs. This intuition is often illustrated with reference to the famous experiment made by Heider and Simmel, at the dawn of so-called 'attribution theory', in which it was argued that people can even attribute beliefs and desires to small geometrical shapes moving in a bidimensional environment, without believing that these figures had genuine mental states [1]. The claim that one may attribute mental states to geometrical shapes, computers, robots, and even living entities without necessarily believing in the existence of these mental states readily accommodates within Dennett's theory of intentional systems. In his [2], Dennett claims that "the definition of intentional systems I have given does not say that intentional systems *really* have beliefs and desires, but that one can explain and predict their behavior by *ascribing* beliefs and desires to them" (p. 91). In his framework, to adopt the intentional stance towards a robot – to ascribe beliefs and desires to it – does not imply believing that the attributed beliefs really exist as such.

While there is a rich and growing literature on people's attribution of mental states to robots (as attested in [3]), few empirical works in HRI research *explicitly* set out to study people's beliefs about the existence of robots' mind. One possible explanation is that this goal is perceived as not particularly important or interesting in the HRI community. This is not surprising, one may say, considering that under a certain interpretation (see, for example, [4]) Dennett himself, who powerfully inspired contemporary research on mental state attribution to robots, held an instrumentalist perspective on the intentional stance. In his perspective, whether or not one believes that the mental states they attribute to the robot exist as such, will not affect the "nature of the calculation" ([2], p. 91) that underlies their predictions and explanations. In other terms, *ceteris paribus*, two agents attributing the very same mental states to the robot, but differing in their ontological commitment, would make the same behavioural predictions and explanations. In a similar vein, Thellman and Ziemke [5] have claimed that "people tend to predict and explain robot behavior with reference to mental states without reflecting on the reality of those states". They also add the following:

> There is to our knowledge no evidence that people's beliefs about the reality of the mental states of robots – or of cartoon characters, thermostats, or fellow humans – affect their disposition or ability to predict behavior. It does not seem to matter, to predict the behavior of an agent, whether the person interpreting the behavior of the agent in question believes that the agent really has mental states.

Here the authors do not argue that people's beliefs about the reality of the ascribed mental states do not affect their ability to predict behaviour, but that there is no evidence pointing to this phenomenon. However, they express the tentative claim that such beliefs do not seem to affect people's predictive processes, and this could be readily interpreted as a reason to eschew the study of the characteristics of the ontological commitments made by people who attribute mental states to robots (even though one may reply that the lack of evidence on a phenomenon is a powerful reason at least to provisionally check if that phenomenon occurs).

The broad goal of this article is to promote a reflection on people's beliefs about the existence of robots' minds. It is suggested here that research on the attribution of mental states to robots should take an 'folk-ontological turn', and explicitly focus on when, why, and for what purpose people make ontological commitments to the mental states they attribute to robots. This broad goal is pursued here via the following, more specific objectives.

The first one is to clarify the notion of 'ontological commitment'. The analysis made here stems from the intuitive understanding of this notion that seems to be presupposed in the HRI literature. It is commonly held there that people may attribute mental states to robots *and* make, or not make, particular ontological commitments to their reality. Ontological commitments, which are typically assumed to be different from mental state attributions, should therefore be conceived as 'things' (possibly, beliefs) that are 'attached to', or conjoined with, mental state attributions. Accordingly, one might attribute to a robot the belief that P and at the same time believe (or not believe) that the belief that P is real. Ontological commitments might be also construed as sorts of modifiers of mental state attributions, or interpretations made by the subject of their very mental state attributions. The lack of an explicit definition of 'ontological commitment' in the HRI literature leaves the door open to these conceptual speculations. Here it will be proposed that ontological commitments are beliefs, but that there is no radical distinction between mental state attributions and people's beliefs about the reality of the robots' mental states. In the perspective defended here – which also hinges upon a plausible analysis of the notion of 'attribution' – different ontological commitments consist in different sets of mental state attributions, and the study of mental state attributions may, *per se*, provide evidence for the study of people's ontological commitments. While this thesis undermines the claim that one can attribute mental states to robots without making any commitments to their reality, it also entails that, even though HRI researchers have so far rarely focused on ontological commitments *explicitly*, they have in fact focused on them at least *indirectly* by studying mental state attributions. This implies that the way is already paved for the 'folk-ontological turn' envisaged before. This objective will be pursued in Section 2.

In the same section, the term 'folk-ontological stance' will be introduced to refer to people's ontological commitments to the mind of robots. Folk-ontological stances towards robots are regarded here as consisting in sets of beliefs concerning the existence of robots' minds. As in 'folk psychology', the term 'folk' suggests that the beliefs at stake are held by humans during ordinary interactions with robots and are not necessarily the result of controlled scientific experimentation. The term 'stance' alludes to the fact that these beliefs constitute a way of thinking about the robot. The term 'ontological' refers to the fact that different folk-ontological stances will differ in whether the subject really believes that the robot has mental states. Note that folk-ontological stances do not concern the reality of the mental states *of the person doing the attribution*. It will be said, for example, that agent A adopts a psychologically realist folk-ontological stance towards a robot if they really believe that *the robots' has certain beliefs*. This stance does not imply any thesis about the reality of A's mental states: simply speaking, it is not about A's mental states, but about A's beliefs about the robot's mental states.

The second objective of this article is to explore the space of the possible folk-ontological stances people may take towards robots. This rich and so far substantially uncharted territory will be modelled taking inspiration from the philosophical debate on scientific realism in psychology and science [6], [7], [8], [9]. The taxonomy of folk-ontological stances sketched here includes psychological realism and its opposite, psychological non-realism. Some varieties of the latter stance, psychological non-realism, are eliminativism (the belief that robots do not possess mental states), agnosticism (the lack of beliefs on the matter), reductionism (the belief that there is something non-mental that makes mentalistic utterances true), fictionalism (the belief that mental states exist in the framework of a fictional story). All these stances will be defined in terms of the beliefs that the subject possesses about the robot. A special place in this taxonomy is occupied by instrumentalism, characterized here not in terms of the possession of beliefs, but in terms of the subject's voluntary decision to postulate mental states to predict and explain the behaviour of the robot. It will be argued that there is an important difference between instrumentalism and all the folk-ontological stances mentioned before. It mirrors the distinction between believing something and *accepting* something [10], a distinction that can be appreciated by observing that people may provisionally accept a premise without believing its truth, just for the sake of the argument. Accordingly, instrumentalism will not be regarded as a folk-ontological stance, but as a folk-epistemological one.

The third aim of this article is to justify the usefulness of the distinctions and conceptual clarifications offered here in empirical HRI research. It will be argued that the analysis of people's folk-ontological stances may increase our understanding of people's perception and understanding of robots. It may reveal characteristics of people's mental models of robots, and of their explanation of robotic behaviour, that have not been explicitly and thoroughly explored so far by the research community. Moreover, it will be argued that people's folk-ontological stances may affect their perception of robots' psychological human likeness. Whether the mind of a robot is understood by agent A as similar to the human mind, is a question to be addressed also by determining whether or not A takes the same folk-ontological stance towards the robot and towards human beings. For example, if A takes a psychologically realist folk-ontological stance toward human beings but conceives the mind of the robot in the non-realist, reductionist way, then it is plausible that the robot will be understood as less human-like than B, who instead takes a psychologically realist folk-ontological stance towards robots and humans. This claim will be refined in section 4, but the gist is that people's ontological conceptions of robots' mind, and of the human mind, make the difference on whether they understand robots' mind as human-like or not. A rich and growing literature supports the thesis that people's perception of human likeness affect the dynamics of their interaction with robots. From the analysis carried out here, it follows that people's folk-ontological stances may affect human-robot interaction too. This is a cogent reason for HRI researchers to take the 'folk-ontological turn' proposed here.

Admittedly, this article does not present any novel empirical or technological result. It offers a philosophical and conceptual reflection on people's ontological commitments to the mind of robots. Still, for the reasons expressed in the previous paragraph, this reflection may be of some interest also for the more empirically oriented HRI researchers. After all, the study of mental state attributions immediately gives rise to philosophical questions that HRI researchers are occasionally happy to address (see, for example, the aforementioned [5]), as they can orient empirical research in important ways. This article intends to contribute to the epistemological debate on people's understanding of the mind of robots and to convince at least part of the HRI community that an 'folk-ontological turn' in the experimental study of mental state attribution can advance research in the field.

Before proceeding, it is worth insisting a little more on what this article is *not* about. Even though mental state attributions are seen here as psychological phenomena, as they are linked to the holding of particular beliefs in the mind of the human agent, this study does not concern the (mental, neural) mechanisms governing the formation and processing of these beliefs. Nor does it concern the determinants or the consequences of the adoption of particular folk-ontological stances. These are subjects of empirical research that may be eventually carried out within the philosophical framework offered here. Some considerations made here resonate with Seibt's reflections on the ontology of social interaction [11], but the goals of the two papers, and the use made of the term 'ontology', are different. Seibt's article illuminates the issue of what social interaction is and uses her result to argue that human-robot interactions cannot be treated as fictional social interactions. Moreover, she offers a classification of forms of human-robot sociality. Even though there may be connections between Seibt's and this paper, the goals pursued here are different, and the study of the ontological commitments that lie behind people's mental state attributions is out of the scope of her work. To the best of the author's knowledge, this is the first paper explicitly arguing that the study of people's folk-ontological stances towards robots may be highly relevant in HRI research.

For reasons that will hopefully be clear in a while, the analysis of folk-ontological stances towards robots cannot proceed without clarifying the notion of 'mental state attribution'. Section 2 is devoted to this goal.

## 2. The truth-maker of mental state attributions

### 2.1. What is an 'attribution'?

According to the comprehensive review made by Thellman and colleagues [3], several different terms are used in the contemporary scientific literature on HRI to address the phenomenon of mental state attribution to robots. They include 'mind perception', 'robot mentalizing', 'theory of mind' (of robots), 'intentional stance', 'mind reading', 'folk psychology', 'anthropomorphism' and, unsurprisingly, 'attribution' of a mind to robots. These notions overlap to some extent, and the authors of the survey suggest that, in the HRI literature, they are all used to refer to the same phenomenon. They recommend that the most intuitive term, 'attribution' (of a mind, mental states, mental capacities), be used, and this recommendation will be accepted in the rest of this paper. This term has quite a long history in the literature on cognitive and social psychology. It is frequently used in the literature on Dennett's intentional systems theory and is the key term in the so-called 'attribution theory', which originated from Heider's psychology of interpersonal relations [12], and was later developed by scholars such as Jones and Davis [13], Kelley [14], and Malle [15]. As noted by Malle in [16], the object of the attribution has changed in this literature: whereas most authors initially developed models of attribution of traits and stable dispositional properties to humans, other scholars now use this term to refer to the attribution of mental states to other agents, which is the use that is typically made of this term in the contemporary HRI literature. The term 'attribution' is also widely used in the literature on the development and the exercise of the so-called theory of mind [17].

Whereas a wealth of studies have been carried out on the determinants of mental state attribution and the mechanisms underpinning it (see [18], [19], [20], and the whole literature on the so-called 'theory theory' and simulationist models discussed in [17]), as well as on how mental state attribution affects social interaction (the 'attributional' theories as Kelley and Michela [21] call them), the very term 'attribution' is typically used as primitive. At least in the relatively circumscribed field of HRI, it is used without any explicit definition (see, for example, [3], [22], [23], [24], [25]). In particular, one question is seldom, if never, directly addressed: what makes it true that agent A attributed a certain mental state to R? In a certain sense, this is a question about what mental state attributions consist in. It can be rephrased in the following terms: what in the world 'out there' must happen, for one to be justified in asserting that A attributed a certain mental state to B? What do mental state attributions correspond to, in the world 'out there'?

As pointed out before, the answer can hardly be found in the literature. In one of the few explicit attempts to define the term 'mental attribution', Brüne and colleagues [26] state that "the term 'mental state attribution' has been introduced to describe the cognitive capacity to reflect upon one's own and other person's mental states such as beliefs, desires, feelings and intentions". This statement is of little help in addressing the problem of attribution truth-makers. What makes it true that John attributed to a robot a certain belief, e.g., that a particular object is a toy horse? Following Brüne and colleagues, one may answer that the truth-maker is John's possession of the cognitive capacity to reflect upon his own and the robot's mental states such as beliefs, desires, feelings and intentions. This answer is unsatisfactory, however, because the truth-maker question (the possession of that capacity) is content-neutral. The same state of affairs in the world 'out there' (John's possession of that capacity) may also make it true that John attributes to the robot the belief that the object is a toy zebra, or the desire to kill John. One might therefore try and build a less content-neutral version of Brüne and colleagues' view: what makes it true that John attributes to the robot the belief that this is an apple is that John possesses the cognitive capacity to reflect upon the robot's belief that this is an apple. This suggestion is more content-specific, but still unsatisfactory for several reasons. John's attribution is a relatively volatile phenomenon. It may be the case that today John attributes to the robot a particular belief, while yesterday, or a minute ago, he might have attributed to him a different belief. The possession of a cognitive capacity is plausibly, instead, a more permanent trait of John's. It is commonly taken for granted, in cognitive science, that cognitive capacities – whatever they are – can develop and deteriorate, but not at the same pace as attributions. It is true that, as stated by Brüne and colleagues, the term 'mental state attribution' is *used* to theorise about people's capacity to reflect upon one's own and other persons' mental states. Still, they do not offer any account of what makes it true that agent A attributes a certain mental state to robot R.[1] To the best of the author's knowledge, no account of attribution truth-makers cannot be found elsewhere in the HRI literature.

Another possible answer is that the truth-makers of mental state attributions must be found in people's exercise of verbal or non-verbal behaviour. This answer has some evident limitations, however. Consider verbal behaviour first. It is true that, in the literature, people's mental state attributions to robots are often experimentally detected by analysing their verbal discourse or using questionnaires in which the participants are asked to choose statements from a list [23], [27], [28], [29]. So, one concludes that John attributes to the robot the belief that this is an apple because he utters the sentence "The robot believes that this is an apple" or because he marks the sentence "The robot believes that this is an apple" in a questionnaire. However, people may attribute mental states to robots also without uttering the corresponding sentence.[2] For this reason, it cannot be the case that what makes it true that John attributes a mental state to a robot is that John utters the corresponding sentence verbally or that it chooses it in a questionnaire. This would be too restrictive a view.[3] Similar considerations could be made about the thesis that what makes it true that John attributes belief B to the robot is that John produces a certain non-verbal behaviour (for example, that he displays certain reaction times when presented with certain stimuli). Mental state attributions need not be accompanied by particular behaviours. Even though this observation – that the mark of attributions cannot consist in particular verbal and motor beahviours – will appear undoubtedly obvious to HRI researchers, the common usage of the term 'attribution' may well generate this kind of misunderstanding. Often, in HRI studies, people's attributions are too directly, and seemingly unproblematically, inferred by their utterances or choices in questionnaires.

---

[1] The notion of mental state attribution, as well as the taking of an intentional stance, are often equated with the adoption of a particular predictive and explanatory strategy. So, for example, Marchesi and colleagues [23] point out that "Adopting the intentional stance refers … to *adopting a strategy* in predicting and explaining others' behavior with reference to mental states". Quoting Dennett, they equate the intentional stance with "the ascription of beliefs, desires, intentions and, more broadly, mental states to a system, in order to explain and predict its behavior". This suggestion offers no easy answer to the problem of attribution truth-makers. What makes it true that John attributes to a robot the belief that that object is a toy horse? Building on the view presented here, one may suggest that the truth-maker of John's attribution consists in the adoption of a predictive and explanatory strategy that refers to the robots' belief that that object is a toy horse. What must be true in the world 'out there', for one to be justified in affirming that John makes this attribution, is that John adopts that strategy. This begs the question of what makes it true that John adopts a particular predictive and explanatory strategy when interacting with the robot. This question may admit of a few possible tentative answers, whose scrutiny is postponed to further analyses. As a general consideration, however, 'reducing' the problem of attribution truth-makers to the problem of strategy-adoption truth-makers ends up increasing the complexity of the issue addressed here, as the latter problem does not seem to be easier to solve than the former one. For this reason, a different solution will be preferred here, as clarified in this section.

[2] Moreover, as pointed out in [30], people may make very different verbal attributions regarding the same robot depending on the way the attribution is elicited. The same person may use mentalistic terms to talk about the robot in spontaneous reactions, and verbally deny that the robot has a mind when carefully reflecting on it.

[3] There is still another reason to exclude this view. The utterance of a mentalistic sentence is a brief phenomenon. It is implausible that what makes it true that John attributes to the robot a certain belief is that John utters the corresponding sentence, i.e., that mental state attribution is true *only while* John is talking. One may respond to this objection by proposing that what makes it true that John attributes to NAO the belief that that object is a toy horse is that John, if asked "What does the robot believe?", would answer "That that object is a toy horse" – or, borrowing from the philosophical jargon, that the truth-maker of John's attribution is a *behavioural disposition* of John's. This possibility will not be explored here, because it gives rise to the vexed problem of understanding what makes it true that somebody or something has a behavioural disposition (for a discussion, see [31]). The point of view proposed in this section partially sidesteps this problem.

A more sophisticated version of these views is that the truth-maker of people's mental state attributions can be identified with the way they treat the robot. It is commonly claimed that people occasionally treat robots *as if* they possessed mental states. Accordingly, one may suggest that what makes it true that John attributes belief B to the robot is that John treats the robot as if it believed B. An instance of this scheme might be: what makes it true that John attributes to the robot the belief that it wants to kill him is that John *treats the robot as if it wanted to kill him*, e.g., he runs away from it or yells "The robot wants to kill me!". This proposal raises the problem of defining what it means that John treats the robot as if it believes that B. This term – to *treat* something as if it had mental states – is typically used to make sense of people's overt behaviour. An external observer sees John run away from the robot and hypothesizes that he is treating the robot as if it wanted to kill him. This circumstance can be more precisely described as follows: to the external observer, John's behaviour *can be best explained* by hypothesizing that he believes that the robot wants to kill him. Putting these considerations together, according to the proposal discussed here, what makes it true that John attributed belief B to the robot is that John's behaviour can be best explained by hypothesizing that he believes that the robot believes B. This proposal suffers from the problems discussed in the previous paragraph: John may attribute belief B to the robot standing still and silent. There is an additional problem though: the truth-maker of John's attribution is *the existence of a theory that best explains* his behaviour. Following this proposal, the problem of attribution truth-makers raises other challenging conceptual problems, widely discussed in the philosophical literature, namely, what it means that a theory *best explains* behaviour and that a theory *exists*. One may well wonder if there are simpler solutions to the attribution truth-makers problem.

*2.2. Mental state attributions as beliefs*

This article takes the point of view according to which the truth-makers of mental state attributions must be sought in the *beliefs of the person doing the attribution*. In this perspective, what makes it true that John attributed a certain mental state to the robot is that John holds a particular belief, or set thereof, about the robot (whose content will be discussed shortly). To identify the truth-makers of John's mental state attributions, one must shift the focus from the robot's (attributed) mental states to John's own beliefs. The fact that John's beliefs, unlike their observable behaviours, are hard to determine does not undermine the plausibility of this suggestion: the question at stake is what mental state attributions consist in, not how they can be studied. John's beliefs about the robot may influence his behaviour, but the same belief can contribute to the production of different observable behaviours in different circumstances, depending on a high number of auxiliary factors, including the content of the other beliefs of John's. Notably, John may attribute a certain mental state to the robot by standing quiet and still: he neither always need to verbally express his beliefs nor express them in the same way.

This perspective resonates with the epistemological presuppositions of much contemporary HRI research on mental state attribution. Indeed, many scholars have claimed that the dynamic of human-robot interaction is affected by people's *mental models* of robots (e.g., [32], [33]).[4] For example, Thellman and colleagues [3] claim that studying people's "mental state ascriptions, i.e., … *people's understanding or mental models of robots* as agents with particular (ascribed) mental states and capacities" (emphasis added) may enable one to address one of the grand challenges of social robotics, that is, to understand "how mental state attributions affect how people interact with robots". Mental state ascriptions are taken in this passage as consisting in people's mental models of robots. Epley and colleagues [19] refer to people's 'anthropomorphic beliefs' in their discussion of how people 'see' robots. Even though these authors do not explicitly define their conception of a mental model, they clearly do not reduce mental state attributions to verbal utterances only. Conceiving mental models as sets of beliefs is admittedly a theoretical choice that could be questioned. However, it is a plausible choice at least to start with. See Figure 1 for a graphical rendering of this idea.

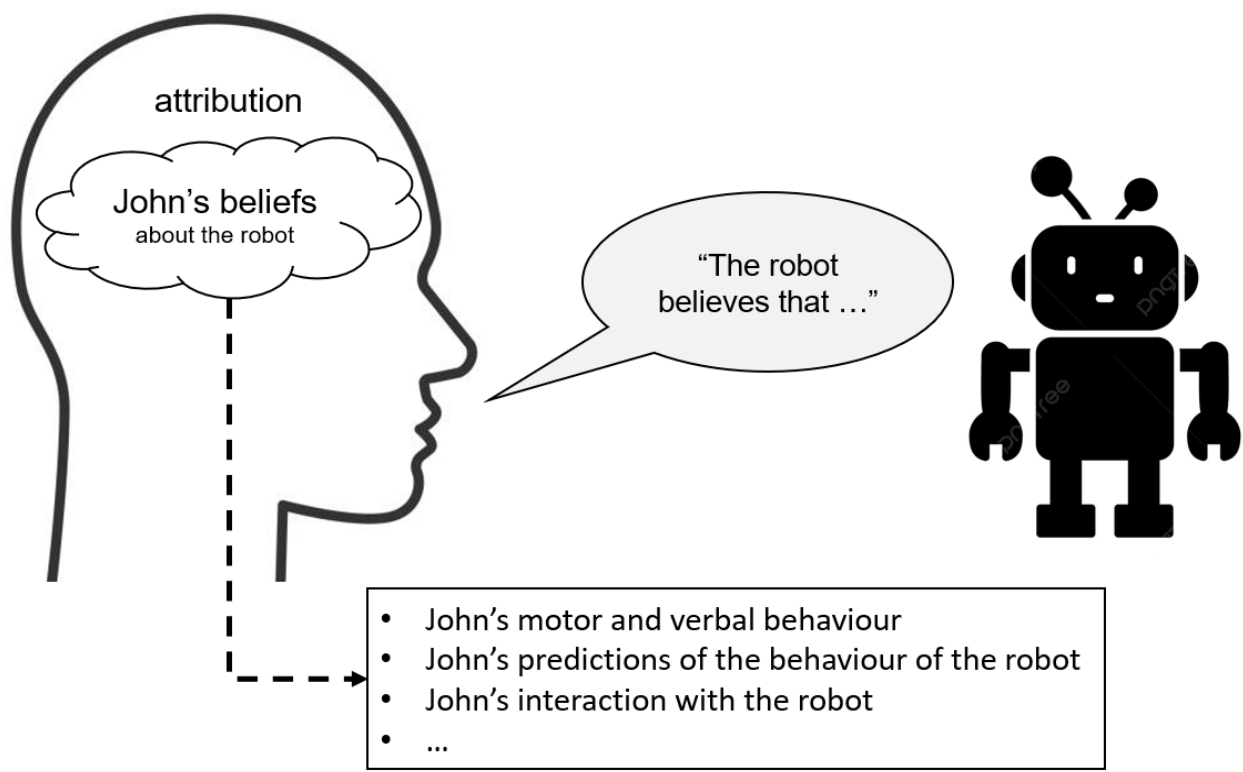

Figure 1 – John attributes a mental state to the robot if they have a certain belief about the robot. John's attribution can shape John's verbal and motor behaviour, his behavioural predictions, and the dynamics of his interaction with the robot.

The discussion has so far been focused on *mental state* attributions, but can be generalized to the attribution of states, properties, mechanisms to a particular system or agent in the following terms:

---

[4] Mental models of robots may not consist in systems of beliefs about the robot. They might consist in mental representations that cannot be properly regarded as beliefs as they are typically conceived in folk psychology. More generally, studying John's mind requires one to develop a theory about it, and it is not obvious that such a theory must be couched in terms of beliefs. The discussion made in this paper is therefore restricted to human mental models (of robots' minds) that are constituted by sets of beliefs about it.

> (ATT) Subject A attributes state / property / mechanism X to agent B if and only if A believes that agent B is in state X / has property X / realizes mechanism X.[5]

Consistently with ATT, from now on, the term 'attribution' will be used to denote a belief in the mind of the person doing the attribution, and the verb 'to attribute' will be used as in the following example: the sentence "John attributes to the robot the belief that today is raining" states that John possesses the belief that the robot believes that today is raining. It is worth stressing that John may also attribute a non-mental state, property, or mechanism to B, and in this case John's belief will not be about B's beliefs. For example, if John attributes to B the property of having a transistor inside, then John believes that B has a transistor inside.

To sum up. The term 'mental model' will be taken here to denote a set of beliefs (or attributions). The fact that John holds a mental model of a robot is interpreted here as the fact that John possesses a set $B = (b_1, \ldots b_n)$ of beliefs about the robot, and their mental model corresponds to those beliefs. Some of these beliefs may have a content that refers to the robot's mind (as in "John believes that the robot believes that today is raining"), while in other cases the content may refer to non-mental properties, states, mechanisms of the robot ("John believes that the robot has a transistor"). John's attributions constitute their mental model.

## 3. Folk-ontological stances

*3.1. What is an* ontological commitment *to the reality of robots' mental states?*

On the one hand, the claim that mental state attributions consist in beliefs possessed by the person doing the attribution might sound relatively unproblematic. On the other hand, it is not *prima facie* obvious how this claim can be reconciled with the view, held by some HRI scholars, that "a person might attribute the behavior of a robot to mental states without necessarily committing to any ontological position about the reality of those mental states" [5] and that "mental state ascriptions *do not necessarily involve any ontological commitments* (i.e., they do not entail beliefs about whether ascribed states are real or fictive)".

The problem at stake is what committing to an ontological position (more briefly, making an ontological commitment) about the reality of a robot's mental state consists in. Since it is a position, it is reasonable to identify it with one or more *beliefs* held by the person making the commitment – as the last quoted statement of the previous paragraph seems to imply. In other words, John's making an ontological commitment to the reality of a robot's mind is to be identified with John's possession of particular beliefs about the robot's mind. What is the content of those beliefs? In section 3 different kinds of ontological positions about the reality of robots' mental states, there called folk-ontological stances, will be identified. Here, to introduce the problem, it may be appropriate to discuss an ontological position that is often implicitly referred to when it is said that somebody makes an ontological commitment to a robot's mental states, i.e., the *psychologically realist* position, according to which one believes that the robot really has genuine mental states. Thus, suppose that John makes a psychologically realist ontological commitment to the robot's being in a certain mental state X. What does John believe about the robot, in this case? That the robot, simply speaking and literally, is in mental state X. Or, that the robot *really is* in mental state X, or that mental state X is *genuinely* such – but it is not clear what difference the words in italic make to the point made here: a simple way to conceive John's psychologically realist position is to identify it with John's belief that the robot *has* mental state X.

---

[5] The notions of state, property, and realization will not be discussed here, to keep the article focused on the objectives stated in the Introduction. It is assumed that all the most influential philosophical analyses of these notions are compatible with the claims defended here.

It follows from ATT that, if John makes a psychologically realist ontological commitment to the robot's being in mental state X, then John attributes mental state X to the robot (and *vice versa*). Indeed, both claims are equivalent to the claim that John believes that the robot has mental state X. Assuming ATT, the attribution of a mental state implies a psychologically realist ontological commitment to that mental state. In this perspective, it is not clear how "a person might attribute the behavior of a robot to mental states without necessarily committing to any ontological position about the reality of those mental states". Mental state attributions are inherently ontologically binding, unless, of course, one drops ATT and analyses attributions in different terms – an analysis that, as pointed out before, has not been offered elsewhere.

These observations can be brought to bear on the distinction made by Thellman and Ziemke [5] between the 'belief question' and the 'attribution question' in HRI research. The belief question concern "people's views on the reality of mental states of robots" and can be formulated as "Do people think that robots have minds?". The attribution question, instead, is: "What kinds of mental states do people ascribe to robots?". The authors are not explicit as to what they mean with 'kinds of mental states', but it is clear from their discussion that, for example, beliefs, or beliefs with a particular content, are kinds of mental states. Under this interpretation, answers to the attribution question will have the form 'Agent A attributes certain beliefs to the robot', or 'Agent A attribute the belief that C to the robot', where C is a proposition. Assuming ATT, these attributions imply a psychologically realist folk-ontological stance, consisting in A's belief that the robot has belief, or that it has the belief that C. But if agent A takes a psychologically realist folk-ontological stance towards the robot, they also believe that the robot has a mind, which is one of the possible answers to the belief question. In the perspective proposed here, therefore, it is not clear how the belief question can be distinguished from the attribution question. To disentangle the two questions, one needs to endorse a philosophical analysis of the concept of mental state attribution that departs from ATT. Since the distinction is clearly important in the perspective advocated by Thellman, Ziemke and other HRI scholars, as it can speak of cases such as Heider and Simmel's famous experiment [1], it is suggested here that a philosophical debate on it should be opened. While waiting for it, ATT will be assumed in the rest of this article.[6]

---

[6] In a few cases, however, it is explicitly suggested that mental state attributions consist in beliefs about the robot. For example, Wiese and colleagues [34] states that "reasoning about the internal states of others is referred to as mentalizing, and presupposes that our social partners *are believed to have a mind*" (emphasis added). In [35], the term

Psychological realism and the various forms of psychological non-realism that will be discussed in the next section are called here folk-ontological stances (in this example, towards the robot) for the reasons anticipated in the Introduction. They are *stances* because they consist in sets of 'background' beliefs that John has about the robot and that can influence his behaviour towards it. They are *ontological* because different folk-ontological stances will differ in whether the subject really believes that the robot has mental states. They are *folk* because they are not the result of philosophical or scientific argumentation carried out on the results of the experimental analysis of the robot's behaviour. 'Folk ontology', here, is used to refer to the ontological dimension of John's folk psychology about the robot. These folk-ontological stances differ from one another in the set of beliefs B possessed by John about the robot.

To sum up. It has suggested here that John's ontological commitments to the reality of the robot's mental states are to be identified with particular beliefs held by John about the robot. In particular, a psychologically realist ontological commitment consists in the belief that the robot has mental states. According to ATT, this is also true when John attributes mental states to the robot. Therefore, in this perspective, it is not clear how the 'belief question' and the 'attribution question' can be disentangled. It is not clear how one can attribute mental states to robots *and* not make any ontological commitment to the robot's mind or make a non-realist ontological commitment. While it is true that John's *saying* "This robot believes that it's raining" does not imply any beliefs about the reality of the robot's mind (because utterances are not reliable indicators of belief), John's *attributing* the belief that it's raining to the robot does imply that John takes a psychologically realist folk-ontological stance. The next question to be addressed is what different kinds of folk-ontological stances towards the robot John might take, and how they relate to John's attributions. This question will be addressed in the following subsections. These observations will be then brought to bear on the study of the dynamics of HRI, with a particular focus on psychological human likeness.

*3.2. Folk-ontological stances: psychological realism and non-realism*

To proceed towards a taxonomy of possible folk-ontological stances towards robots, it will be useful to contrast psychological realism with psychological non-realism. The former stance has already been introduced, but it will be discussed here in a more explicit way with reference to mentalistic utterances. Suppose that John says, "NAO believes that this is an apple". For short, let B refer to the set of beliefs held by John in this circumstance (also known as his mental model of the robot), and F be the proposition "this is an apple".[7] John takes the folk-ontological stance called 'psychological realism' if he believes that NAO believes that F, or equivalently, if B includes the belief that NAO believes that F. If the content of John's belief is that NAO believes that F, John believes that NAO *has such a belief*, i.e., that NAO's belief *exists as such*. Note that what is discussed here is not John's utterance but his inner belief. It can be taken for granted

---

'mind perception' is used to denote "the belief that social cues originate from an entity with a mind, capable of having internal states like emotions or intentions".

[7] No restriction is made here on the characteristics of the robot. Whether some kinds of robots tend to elicit some folk-ontological stances and not others is a question for future research, which goes out of the scope of this article.

that John might say "NAO believes that this is an apple" without really intending to assert that NAO has a genuine belief. There is a relatively clear sense in which John may *say* that NAO believes that F *in a nonliteral sense*. The utterance ought to be nonliterally interpreted because John, in this case, would not really believe that NAO believes F: he would probably hold different beliefs about NAO. But it is not clear how John might *believe* that NAO believes that F *in a nonliteral sense*. Unlike utterances, beliefs are not things that can be had nonliterally. If John believes that NAO believes that F, then John believes that NAO's belief exists. To say that a robot has a belief, and to believe that a robot has a belief, are clearly different things, and while the former case does not imply psychological realism (because utterances can be pronounced nonliterally), it is not clear how the latter could *not* imply psychological realism.[8]

Psychological realism, as conceived here, is cognate of scientific realism in psychology. Scientific realism in psychology is the epistemological and ontological position according to which mature psychological theories are literally true: it consists in the thesis that the mental entities and properties that these theories postulate actually exist (forms of psychological realism are discussed in [36] and [6]). While there is a clear connection between scientific realism in psychology and psychological realism as a folk-ontological stance, the two are not claimed here to coincide, at least at a psychological level of analysis. A first reason is that scientific realism in psychology is a view that scientists and philosophers endorse concerning the theoretical entities posited by scientific theories about the mind, while the stance discussed here is taken in non-scientific, ordinary interactions with the external world. A second reason is that epistemological and ontological positions are consciously and deliberately *accepted*, and there is a clear sense in which one can accept a philosophical thesis without really believing it. If John sees a dog in his living room, he will believe that there is a dog in his living room. However, John also may eventually accept that he is hallucinating, because this is what a doctor and his best friend are telling him, and because he remembers that the day before he drunk too much. Belief and acceptance can coexist: there is a sense in which John may continue to believe that a dog is in his living room, but at the same time decide to go to the hospital to recover from his delusional state. As a more mundane example, people used to believe that the earth was at the centre of the universe. Eventually, strong arguments were developed for a very different thesis. Plausibly, there has been a time in which even those who produced those arguments experienced a conflict between what they involuntarily believed and the new thesis that they had to accept. Eventually, people changed their beliefs about the position of the earth in the universe. Acceptance may produce belief change (and one's beliefs may shape the process of acceptance), but this observation does not undermine the distinction between belief and acceptance, a distinction that has been explored in a long tradition of philosophical research (see, for example, [10], [37]).

In this perspective, scientific realism in psychology is an epistemological and ontological thesis that some people accept, while folk-ontological stances are sets of beliefs that people may possess. The two can influence each other without coinciding. The distinction between belief and acceptance will be used in the next subsections to characterize the agnostic and the instrumentalist folk-ontological stances toward robots.

---

[8] It will be assumed here that, if B includes this belief, then John also believes that NAO has a mind, insofar as to have a belief implies being in a particular state of mind.

The claim that, if John attributes a mental state X to NAO, then John makes a psychologically realistic ontological commitment does not clearly imply that psychological realism is the only possible folk-ontological stance John can take. It only implies that, if John does not take a psychologically realistic folk-ontological stance in this case, then he does not attribute mental state X to the robot (he does not believe that the robot has mental state X). But he could attribute other kinds of states to the robot. Or, he could attribute no mental states whatsoever to it. There is a wide spectrum of possible alternatives between psychological realism and the making of no ontological commitments. All these may result in the same utterance "NAO believes that this is an apple", yet, as it will be shown in the following sections, some of them might make the difference in John's perception of NAO's psychological human likeness.

To illustrate these non-realist positions, it is useful to elaborate on the distinction mentioned above. What are the alternatives to psychological realism? At a first glance, the following options can be envisaged:

1) John makes no ontological commitment whatsoever to the reality of the robot's mind, which corresponds to having no beliefs about its mind. This case will be called *agnosticism*, a condition that can be accompanied by *instrumentalism*, and will be discussed in section 3.4.

2) John makes an ontological commitment that is different from psychological realism. He believes that the robot is in some non-mental state that, from a theoretical point of view, can be nevertheless regarded as 'mental' under a particular interpretation of the term. This will correspond to folk-ontological stances called *eliminativism*, *reductionism,* and *fictionalism* (section 3.3).

These two circumstances have something in common, namely, the fact that they are alternative to psychological realism: John's knowledge base about, or mental model of, the robot does not include the belief that NAO believes that F. This will be called *psychological non-realism*. One may be classified as psychologically non-realist as far as the belief that F is concerned, i.e., not believe that the robot believes that F. Or, they can be regarded psychologically non-realist in regards to a wide range of possible beliefs held by the robot, e.g., not believe that the robot has belies whatsoever.

But while in case 1 (corresponding to agnosticism and instrumentalism) John's mental model does not include any belief concerning NAO's mind, in case 2 (eliminativism, reductionism, and fictionalism) John's mental model includes beliefs about NAO's being in certain non-mental states that can be regarded as mental under a particular interpretation of the term. While in the first case John is ontologically noncommittal about the robot's mind, in the second case John makes ontological commitments that are different from psychological realism. In the terminology proposed here, psychological non-realism is a broad class of possible folk-ontological stances that encompasses cases of no ontological commitment (agnosticism and instrumentalism) and cases of non-psychologically-realist ontological commitment (eliminativism, reductionism, fictionalism). While a different terminology might be adopted, it is suggested that the folk-ontological stances identified here may constitute a useful taxonomy to understand the dynamics of HRI and in particular the phenomenon of human likeness.

*3.3. Folk-ontological stances: eliminativism, reductionism, fictionalism*

Suppose that John's mental model of the robot (a) *does not include* the belief that NAO believes that F and (b) includes the belief that NAO *does not believe* that F. In this case, John is a *psychological eliminativist* about NAO's possession of that specific belief. One may be eliminativist towards a specific belief of NAO's (i.e., one might simply believe that NAO does not have that specific belief, but can have other beliefs), or one might take a more general eliminativist stance, believing that NAO cannot have beliefs, that it does not have a mind. The more interesting case, for the present discussion, is the latter one – John's belief that NAO does not have a mind. What matters, here, is that John may take a folk-ontological psychological eliminativist stance that consists in having particular beliefs about the robot, namely, that it does not have a mind, and, consequently, mental states of any sort (for an eliminativist position about folk psychology, see, e.g., [38]).

How could John utter the sentence "NAO believes that this is an apple" and be a psychological eliminativist? An easy answer is that John's utterances about the robot do not depend on his beliefs about the robot only, but also on a variety of internal and external contextual factors. So that utterance may be caused by beliefs and desires that do not concern NAO at all, for example, by the desire to instil a certain idea in Dennis, a third observer. A more interesting possibility is that John is not speaking literally, and the set of his beliefs about the robot includes beliefs about the robot that are different from the belief that NAO believes that F. In particular, John might want to assert something *different* from the fact that NAO has a belief, and that the content of this belief is F. In line with Toon [8] and Yablo [9], two additional folk-ontological stances can be identified, that are called here *psychological reductionism* and *psychological fictionalism*.

If John's folk-ontological stance is *psychological reductionism*, John does not believe that NAO believes that F. However, John believes that something *else* about the robot is true – something that would make it reasonable to say, "NAO believes that this is an apple" and that would make it unreasonable to say, in the same conditions, "NAO believes that this is a banana". This 'something else' might be, for example, that NAO is in a particular electrical or computational state. To illustrate, suppose that John is a robotic engineer and matured a firm eliminativist folk-ontology about the mind of robots. Robots do not have mental states, he believes. Robots are extremely complicated electronic circuits whose functioning can be described using the language of physics or, at a higher level of abstraction, using the language of computer science. John knows that NAO has an object-detector module whose output may be 'apple' and 'banana'. In the scenario above, assuming that the robot correctly represented his verbal request, John hypothesizes that the output of the object-detection module was 'apple'. He then utters the sentence "NAO believes that that object is an apple". This utterance does not flow from his being a psychological realist: he is not. Neither he is simply expressing an assumption that, instrumentally, could explain NAO's behaviour. Rather, he believes that there is something, in principle describable using the language of computer science or even physics, that would make it reasonable to utter that sentence more than the sentence "NAO believes that this is a banana". His set of beliefs (i) does not include the belief that NAO believes that F, but (ii) includes a belief concerning the physical or computational state of the robot. Psychological reductionism, conceived in this way, is a non-realist stance characterised by the fact that John possesses some beliefs about the non-mental characteristics of the machine.[9]

*Psychological fictionalism* is yet another possible folk-ontological stance towards NAO. Unlike the previous example, suppose that John is not an expert in robotics. However, he wants to play a make-believe game with Anne, a child. They are inventing a story in which NAO is a friend. By saying "NAO believes that this is an apple", John merely wants to assert that, in the fictional story they have invented, NAO believes that this is an apple. Thus, while it is not true that John believes that NAO believes that F, it is true that John believes that *in the fictional story they have invented* NAO believes that F. This folk-ontological stance has been called psychological *prefix* fictionalism (e.g., in [8]) because the clause 'in the fictional story they have invented' is a prefix that, albeit not verbally pronounced by John, when added to the text of John's utterance, defines John's belief. This folk-ontological stance is different from psychological reductionism: here, John is not asserting that something non-mental is true of NAO which makes it reasonable to say, "NAO believes that this is an apple". Rather, he is asserting that some state of affairs occurs in the framework of a fictional story. To understand, compare this case with a make-believe game played by a child and two objects, e.g., a marble and a toy block. In the framework of that fictional game, they may pretend that the toy block is the son, the marble is his mother, and make several mentalistic assertions about these two objects, without believing that the two objects have physical characteristics that make these assertions sensible. This may also be the case with fictional stories. If John says, "Sherlock Holmes lives in Baker Street", he is not asserting that something in the world 'out there' is true which makes it sensible to say that Sherlock Holmes lives in Baker Street. He is expressing a belief, whose content is that, *in Conan Doyle's stories* (this is the prefix), Sherlock Holmes lives in Baker Street. In the robotic scenario considered here, psychological prefix fictionalism is a folk-*ontological* stance towards NAO because it expresses the belief that something – NAO's belief that F – exists in the context of a fictional story (for an extensive philosophical discussion of what it means to assert something in the context of a fictional story, see [41]). Clark and Fischer [42] have recently proposed a fictionalist understanding of human-robot interaction.

To sum up. The spectrum of the possible folk-ontological stances that John may have towards NAO includes psychological realism and psychological non-realism. Psychological non-realism is in fact an umbrella term which encompasses psychological eliminativism, reductionism, and fictionalism. These stances correspond to different kinds of ontological commitments to the reality of NAO's mind.[10]

---

[9] Even though this article does not intend to make any particular claim as to how people's folk-ontological stances affect their prediction of the behaviour of robots, it is worth recalling here that, as pointed out by a number of authors (e.g., [2], [39], [40]), it is rarely possible to predict robot behaviour based on knowledge about the complex physics or the computations that go on inside robotic systems. This does not rule out the possibility that people adopt this FOS in particular circumstances.

*3.4. Agnosticism, instrumentalism, and a folk-epistemological stance*

Another alternative to psychological realism introduced in section 3.2 (point 1) is called *agnosticism*. If John is agnostic, he possesses no beliefs whatsoever about NAO's mind. In particular, his mental model of the robot includes (a) neither the belief that NAO *believes* that F, (b) nor the belief that NAO *does not believe* that F. Psychological agnosticism is a case of psychological non-realism, because of (a), but is different from psychological eliminativism because of (b), as John does not believe that the robot does not believe that F. Agnosticism is compatible with reductionism and fictionalism: John may believe neither that NAO believes that F nor that NAO does not believe that F, and at the same time believe that NAO is in a certain non-mental state that, via reduction or in the fictional way, may be regarded as 'mental'. However, nothing rules out the possibility that John is psychologically agnostic without being reductionist or fictionalist. In this quite radical case, John's knowledge base about the robot would be characterised by a singular absence of beliefs about it. It is worth stressing that agnosticism is not defined in terms of *verbal expressions* of agnosticism or uncertainty. Cases in which a subject says, "I do not know whether robots have a mind" or "I am not sure whether robots have a mind" need not qualify as cases of agnosticism. Utterances *per se* are not necessarily informative about the subject's inner beliefs. And this construal of agnosticism would rely on the questionable assumption that the truth-maker of attributions consist in people's verbal utterance, an assumption that, it is safe to say, few HRI researchers would seriously hold. Therefore, while these utterances are compatible with agnosticism, agnosticism cannot be plausibly characterised as the condition in which one verbally expresses agnosticism or uncertainty. Agnosticism is the condition in which the subject's knowledge base contains no beliefs about whether robots have or do not have mental states, regardless of what they say.

Agnosticism may (but need not) be accompanied by *instrumentalism*. In philosophy of science, the term 'instrumentalism' typically refers the view according to which the theoretical entities postulated by a theory are regarded as explanatory or predictive tools, with no ontological commitments attached. For an instrumentalist concerning psychology, mental terms refer to useful instruments for calculating behavioural predictions or building explanations, but the judgment is suspended about their existence (for a philosophical discussion of instrumentalism in psychology, see [43]). In the scenario considered before, John would be a psychological agnostic *and* instrumentalist if (1) he believed neither that NAO believes that F nor the contrary, and (2) decided to postulate the existence of NAO's beliefs only as tools to explain or predict its behaviour, without making any ontological commitment to their existence. Instrumentalism can explain how agnosticism can be accompanied by the use of mentalistic language to describe and explain robotic behaviour. John may say "NAO believes that this is an apple" and be agnostic and instrumentalist in the sense discussed here. His utterance would be connected to his decision to postulate the existence of beliefs in NAO's mind as tools to explain and predict its behaviour, with no ontological commitment whatsoever.

---

[10] Note, again, that these different folk-ontological stances do not invariably determine particular verbal utterances. For example, the fact that John is a psychological realist does not compel him to pronounce mentalistic discourse about the robot. Conversely, John's verbal utterances are not clear signs of his folk-ontological stance. For example, John may well utter a sentence like "NAO does not have a mind" and yet be psychologically realist: perhaps he wants to deceive the listener, or he is not aware of having psychologically realist beliefs. In the latter case, his beliefs may reveal themselves in (viz. cause) some aspects of John's non-verbal behaviour. Conversely, can John utter the sentence "NAO believes that this is an apple" and still be psychologically eliminativist or agnostic? Surely, this can be the case. In particular, it may be the case that John, with this sentence, does not want to express the belief that NAO believes that F, but he wants to express *other* beliefs about it. These beliefs concur to define other possible folk-ontological stances towards the robot.

The instrumentalist use of mentalistic language is often discussed in the HRI literature. For example, Thellman and colleagues [3] point out that "many people might say that their robot lawnmower wants to avoid colliding with trees, although they would not say it has a mind, a will, or desires. In other words, it is not uncommon to conceptualize the behavior of robots as mind-governed without necessarily believing that robots really have minds, similar to how we interpret the behavior of fictional characters, companies, and nation-states". In the same article, the authors note that "mental state terms are to some extent treated inconsistently across studies as either metaphorical or literal by enclosing them (or not) in quotation marks" and that "mental state ascriptions *do not necessarily involve any ontological commitments* (i.e., they do not entail beliefs about whether ascribed states are real or fictive)]". Notably, as pointed out in the Introduction, Dennett himself presented the intentional stance as a strategy in which the subject treats another system as if it had mental states, without necessarily believing that these mental states exist as such [2]. In their everyday interaction with NAO, one might indeed embrace a combination of agnosticism and instrumentalism and decide to postulate the existence of NAO's belief that F as a fictive state, useful to explain or predict its behaviour, without believing that NAO's belief exists as such.

However, it is far from obvious that agnostic instrumentalism can be conceived as a kind of mental state attribution. Consider agnosticism and instrumentalism separately. Agnosticism, *per se*, is the situation in which John makes no ontological commitment to the reality of NAO's mind. One is agnostic *and* instrumentalist, in the sense discussed here, if they do not make any ontological commitment to the reality of a robot's mind, and at the same time decides to explain and predict its behaviour as if it were generated by mental states and mechanisms. Now, if agnosticism means having no beliefs about a robot's mind (i.e., neither believing that it has a mind, nor that it does not have a mind), it follows from ATT that an agnostic does not attribute any mental state to the robot. For, according to the analysis of mental state attributions discussed before, to attribute a mental state to the robot is to believe that the robot has that mental state, and vice versa. So, if one is agnostic and instrumentalist, the 'agnostic' factor does not correspond to a mental state attribution. What about the other factor, instrumentalism? Does it *per se* consist in, or entails, mental state attribution?

Arguably, not. An agnostic instrumentalist has no belief whatsoever about the robot's mind but *decides* to postulate mental states in the robot only as tools to explain and predict its behaviour. Such a decision resembles more an act of *acceptance* or *presupposition* than the holding of a belief. In theorem proving, one may accept of presuppose the truth of a premise even when they do not believe that it is true, just for the purpose of proving the theorem. Acceptance has been defined by Stalnaker [37] as "a broader concept than belief; it is a generic propositional attitude concept with such notions as presupposing, presuming, postulating, positing, assuming and supposing falling under it. […] To accept a proposition is to treat it as a true proposition in one way or another – to ignore, for the moment at least, the possibility that it is false. […] To accept a proposition is to act, in certain respects, as if one believed it" (see also [44]). Importantly for the present purposes, one may accept a proposition without believing that it is true. Conversely, one may believe that a proposition is true without accepting it, i.e., without acting as if they believed it. Hallucination is a case in point: one may believe that there is a dog in the living room but fail to accept it, as their doctor provides convincing evidence for the contrary. Intellectually honest people often acknowledge that they have built-in racial prejudices (beliefs) that they refuse to accept, or to take as a basis for rational action. The distinction between belief and acceptance is discussed at length by [10]. The claim made here is that agnosticism and instrumentalism radically differ from one another from a psychological point of view. Agnosticism is a matter of beliefs – is the situation in which the subject has no beliefs about the robot's mind, thus makes no mental state attribution. Instrumentalism does not re-introduce mental state attributions – otherwise, it would override agnosticism. It is not a matter of beliefs, but of the more or less conscious decision to assume, presuppose, accept the truth of some propositions – for example, the proposition that NAO believes that F – and act accordingly.

So far, instrumentalism has been discussed in combination with agnosticism. An agnostic instrumentalist makes no ontological commitment to the existence of a robot's mind but accepts (without believing) that the robot has some mental states in order to predict and explain its behaviour. According to the present analysis of attribution and instrumentalism, agnostic instrumentalism does not correspond to a mental state attribution (a position that is hard to reconcile with the aforementioned claim that people may attribute mental states to robots without making any ontological commitments, just for the purpose of explaining and predicting its behaviour). Being compatible with agnosticism, instrumentalism need not imply any ontological commitment. However, *per se*, instrumentalism alone (without the agnostic component) is in principle compatible with all the folk-ontological stances discussed before – even with psychological realism. It may be interesting to determine empirically when people adopt an instrumentalist stance towards the robot, and what ontological commitments they are making when they decide to postulate mental states in the instrumental sense, e.g., for explanatory or predictive purposes. Instrumentalism surely deserves a place in the taxonomy proposed in this article. However, it deserves a separate place, as it is not, strictly speaking, a folk-ontological stance. Folk-ontological stances are characterised by different sets of beliefs about the robot, but instrumentalism, as pointed out here, is not a matter of belief. Here it will be regarded as a *folk-epistemological* stance towards the robot, to distinguish it from the set of folk-ontological stances, and to highlight that it corresponds to a decision specifically meant to facilitate the production of explanations and predictions, and more generally, the acquisition of knowledge about the robot.

*3.5. Summary: a taxonomy of folk-ontological stances (and a folk-epistemological stance) towards robots*

To sum up. People may take several kinds of folk-ontological stances towards the robot they are interacting with, regardless of what they say or do. Folk-ontological beliefs lie 'behind' what they do and say, exactly like all the other beliefs that modulate people's behaviour. Some possible folk-ontological stances towards robots (plus one epistemological stance) have been identified in this section, drawing from the philosophical literature concerning scientific realism. They are summarized in the following table.

Table 1 – A provisional taxonomy of folk-ontological (and a folk-epistemological) stance towards robots

| **Folk-ontological stance** | **Description** |
| --- | --- |
| **Folk-psychological realism** | *The subject believes that the robot has mental states, or a particular mental state P.* |
| **Folk-psychological non-realism** | *The subject does not believe that the robot has mental states, or a particular mental state P.* |
| **Folk-psychological eliminativism** | *The subject believes that the robot does not have mental states, or a particular mental state P.* |
| **Folk-psychological reductionism** | *The subject believes that the robot is in a non-mental (e.g., physical) state that can be reasonably expressed through a mentalistic statement.* |
| **Folk-psychological fictionalism** | *The subject believes that the robot has a mental state in the framework of a fictional story.* |
| **Folk-psychological agnosticism** | *The subject believes neither that the robot can have mental states (or that is in a particular mental state P) nor that the robot cannot have mental states (or that is not in a particular mental state P).* |
| **Folk-psychological instrumentalism** *(Folk-epistemological stance)* | *The subject deliberately accepts that the robot has certain mental states, in the same way as when one accepts a premise only for the sake of a logical argument, in order to explain or predict its behaviour.* |

This taxonomy[11] can be used to formulate some empirical hypotheses about what may be going on in an agent's mind when they interact with robots, or even non-robotic, artificial systems. Consider the following passage, in which Thellman and Ziemke [5] comment on Heider and Simmel's experiment: "Clearly, a person might attribute the behavior of a robot to mental states without necessarily committing to any ontological position about the reality of those mental states. Indeed, people commonly ascribe mental states to cartoon characters and animated geometric figures (Heider & Simmel, 1944)". Preliminary, it may be useful to deploy the previous discussion about the notion of 'attribution' to interpret the first part of this claim. The authors claim that people may attribute mental states to a robot without making any ontological commitments to their reality. As shown before, it is not clear how this can happen. If what they mean is that people may say that a robot has mental state without holding any belief about the existence of these mental states, then the passage is perfectly clear and reasonable; however, this would presuppose that 'to attribute' means 'to say', an evidently controversial position. If, instead, attributions consist in the agent's beliefs (as implied by ATT), then what is implied here is that the agent can believe that the robot has mental states and, at the same time, not believe that it has mental states (i.e., not make any ontological commitments to them). One might reply that what is really meant here is that the agent believes that the robot has mental states, but in a figurative sense. But, while it is perfectly clear that one may *say* things in a nonliteral sense, it is not equally clear what does it mean for somebody to *believe* something nonliterally. Utterances, not beliefs, are things that can be interpreted in a literal or nonliteral sense.

---

[11] Note that the categories identified here only loosely correspond to philosophical options discussed in the scientific realism debate. Some distinctions made in philosophical discussions about instrumentalism and fictionalism in science and psychology have been ignored and the folk-ontological stances presented here lump together interpretive

As to Heider and Simmel, the subject watching the video may adopt several folk-ontological stances, including those identified in this article. The agent may be psychologically realist, believing that the geometrical shapes have mental states. This corresponds (via ATT) to attributing mental states to the shapes. This case is theoretically possible, but clearly implausible. At this point, there may be many possible forms of non-realism. Sure enough, one is them is agnosticism: regardless of what they say on the matter, their mental model of the shapes does not include any beliefs about whether they have mental states or not. The subject might also accept, without believing, that the shapes have mental states in the same way as when one accepts a premise only for the sake of a logical argument. This would be instrumentalism. Recall that agnosticism is characterised by a total absence of beliefs. It is safe to hypothesize that such a *tabula rasa* would be quite improbable; and that the subject's use of a mentalistic language to describe Heider and Simmel's scene (e.g., that the large triangle does not want to marry the circle) is more probably backed by some beliefs about what is behind the behaviour of the shapes. Thus, the subject might be psychologically eliminativist, believing that the shapes do not have mental states. They might also be psychologically reductionist: the shapes do not have mental states but possess non-mental states that make it reasonable to say that the large triangle does not want to marry a circle (and unreasonable to say the opposite). Still another possible folk-ontological stance is fictionalism: the subject believes that the shapes possess mental states in the framework of a fictional story. Intuitively, this is a plausible interpretation of what happens in Heider-

---

categories that have been distinguished in the philosophical literature. Moreover, as pointed out before, the folk-ontological stances introduced here are not claimed to coincide with the corresponding philosophical positions about the reality of the mental entities postulated by mature psychology and cognitive science. Nevertheless, this section offered a framework that can be used as a starting point to elaborate a richer and finer-grained taxonomy, and as a reference to set up empirical studies on people's folk ontological beliefs.

and-Simmel-like scenarios. Note however that fictionalism readily accommodates with eliminativism: in this case, the subject would believe that the shapes do not have mental states, only that they have them in the context of a fictional story (in the same way as when one believes that Sherlock Holmes lives in Baker Street in the framework of Conan Doyle's novels). Since eliminativism is view about the reality of the shapes' mental states, this interpretation entails that the subject *is* adopting an ontological stance towards the mental states of the shapes, and that the same happen when people utter sentences about the "mental states [of] cartoon characters and animated geometric figures".

## 4. Folk ontology and psychological human likeness

The determination of people's understanding of robots is an intellectually interesting goal *per se*, and the taxonomy offered here may be (at least provisionally) helpful to explore the folk-ontological dimensions of people's mentalistic explanations and predictions. All the considerations made here build on a particular analysis of (mental state) attributions as beliefs in the mind of the subject, and on the idea that folk-ontological stances are beliefs too. Unless one provides a different analysis of (mental state) attributions, the distance between the 'belief question' and the 'attribution question' seems to be much shorter than what is claimed in the literature – more specifically, there is no appreciable difference between the two. Folk-ontological stances are not 'attached to' mental state attributions, they are not things that 'wrap' attributions in an ontological interpretation (and can be 'peeled away' to be free from ontological commitments). In the sense discussed here, they consist in attributions to (via ATT, in beliefs about) the robot.

As such, this article may be read as issuing a conceptual or an empirical challenge to HRI researchers. The *conceptual* one is to develop an explication of the concepts of 'mental state attribution' and 'ontological commitment' that, unlike ATT, can support both the distinction between the 'belief question' and the 'attribution question' made before, and the claim that people's ontological commitment to the reality of the robots' mental states cannot affect their predictive abilities (and that only attributions would do, as suggested in the literature). The *empirical* challenge might be taken by those who accept the analysis offered here. If the distance between one's mental state attributions and their beliefs on the reality of these states is shorter than commonly suggested in the literature, it might well be the case that people's folk-ontological stances affect people's predictive and explanatory abilities. In the aforementioned passage, Thellman and Ziemke [5] claim that no evidence has been collected on this matter so far. The analysis offered here may help researchers kick off the 'folk-ontological turn' that has been already introduced in the previous paragraphs.

The rest of this paper will bring the discussion made so far on human likeness. More specifically, it will be suggested here that one's folk-ontological stance towards the robot may affect their perception of robot *psychological* human likeness.

It is widely acknowledged that the degree of perceived robot human likeness may significantly affect various dimensions of human-robot interaction. Ciardo and colleagues [45] found that the quality of collaborative action can be affected by how much the robot is robot human-like. The degree of human likeness has been found by Fortunati and colleagues [46] to affect people's expectations about the emotional and cognitive capacities of robots, which in turn likely affect how people interact with them. In voice conversations, human-like robots are more pleasant, engaging, and likeable, and evoke less negative attitudes than robots with a low degree of human likeness [47]. Human-like robots tend to be perceived as agents, able to control their actions and outcomes, more than non-human-like robots, possibly because it is easier for humans to formulate a sensorimotor representation of their behaviour [48]. As argued in [49], [50], the degree of human likeness can also affect moral judgements. The so-called android science research program [51], [52], [53] is based on the assumption that robots endowed with high degrees of human likeness may be useful to study human social cognition and the dynamic of human-robot coordination. This is only a small part of the literature showing the importance of human likeness in the study of human-robot interaction and the design of interactive robots (see also [54] on this topic). High degrees of human likeness may render the robot uncanny (see [55] for a review) and, for this reason, some authors – including [56] – argue that robots should retain a certain degree of robot-ness and product-ness so that users perceive them as objects to be used and do not develop false expectations about them.

This said, what does it mean for a robot to be human-like? Several dimensions of robot human likeness have been identified in the literature (see, for example, [27], [54], [57], [58], [59]). One of them is psychological: to be human-like may also mean to have a mind that is 'like' the human mind [29], [60]. The many attempts to assess whether, and under what conditions, people attribute mental states to robots (e.g., [61], [62], [63]) may be interpreted as attempts to assess when robots are perceived as human-like from a psychological point of view. And psychological human likeness may be an important determinant of HRI dynamics, as empirically shown, among other studies, in [64], [65].

What does it mean, then, for a robot to have a human-like *mind* for an external observer? One plausible answer is that the psychological human likeness of robot R for agent A depends, among other factors, on the difference between A's mental model of the robot's mind and A's mental model of the human mind. People's mental models of robots' mind may vary depending on several factors that have been widely studied in the literature (the robot's physical appearance and motion, as well as people's internal motivations as in Epley's three-factor theory, [19]). And people may also have different mental models of the human mind. They might also display various forms of *dehumanization*, in the sense discussed by [66], "whereby people *fail* to attribute humanlike capacities to other humans and treat them like nonhuman animals or objects" (p. 59). The thesis proposed here is that, since people's folk-ontological stances towards robots (and humans) are integral part of their mental models of robots (and humans), they may significantly affect perceptions of robot psychological human likeness. And, since perceptions of robot human likeness may affect the dynamics of HRI as recalled before, the study of people's folk-ontological stances towards robots may illuminate some aspects of people's interaction with robots.

The idea that folk-ontological stances are integral part of people's mental models of robots has been discussed before. People form their mental models of a robot by attributing states, properties, and mechanisms to it. Via ATT, this corresponds to forming beliefs about its states, properties, and mechanisms. Some of these beliefs (or attributions) will concur to the formation of a folk-ontological stance, in the way discussed in section 3. When one forms a folk-ontological stance towards a robot, the latter is integral part of their mental model of, or set of beliefs about, the robot. The same holds for people's mental models of other human beings. So, a number of interesting cases can be envisaged.

For example, consider the two following cases: 1) John adopts a psychologically realist folk-ontological stance both towards humans and NAO. He attributes mental states to other humans and to NAO, in the sense of 'attribution' discussed before. 2) John adopts a psychologically realist folk-ontological stance towards humans, but a psychologically non-realist stance towards to NAO. Sure enough, in condition 1, John's mental models of NAO *could* be very different from his mental model of his fellow humans: he may, e.g., attribute mental states to NAO and to human beings that differ in the content. For example, he may believe that NAO wants to recharge his battery, and never attribute this belief to human beings. However, what is claimed here is not that people's adoption of the same folk-ontological stance towards robots and human beings will be reflected in the very same mental model of the two kinds of systems, but rather that, if one's folk-ontological stance towards humans is different from their folk-ontological stance towards robots, they will not perceive the robot as human-like. To illustrate it may be useful to consider different sub-cases of condition 2.

Suppose that John, having a psychologically realist stance towards human beings, is *eliminativist* as far as robots are concerned. In this case, his mental models of the robot and of human beings will greatly differ from one another not only in the content of the ascribed beliefs. While John believes that humans have mental states, he will believe that robots do *not* have mental states. If to be psychologically human-like consists in having a mind that is 'like' the human mind, John will not perceive the robot as human-like from a psychological point of view. It is reasonable to believe that John's folk-ontological stances (towards humans and robots) will affect, e.g., his answers to questionnaires to measure anthropomorphism [27] and his moral judgments about the robot's actions. Now consider other two options discussed in the previous section, reductionism and fictionalism. John is *reductionist* towards the robot if he believes that the robot does not have mental states, but that it has non-mental states (e.g., physical) that could be reasonably expressed in mentalistic terms. As far as human likeness is concerned, this case is not that different from eliminativism: John will not perceive the robot as having a mind similar to the human mind (provided, as initially assumed, that he is psychologically realist towards human beings). Consider also *fictionalism*. John is fictionalist if he believes that the robot has mental states in the framework of a fictional story, in the sense discussed before. Since, by assumption, he adopts a very different stance towards human beings, he will perceive the robot as non-human-like.[12]

Instrumentalism has been dubbed a 'folk-epistemological stance' in the previous section. It has been suggested that John is instrumentalist if he deliberately accepts certain assumptions about the robot for explanatory or predictive purposes, in the same sense in which one deliberately makes assumptions in theorem proving or counterfactual reasoning. It has also been suggested that people need not believe what they accept (as is typically the case in counterfactual reasoning) and that the converse is also true, they need not accept what they believe. Therefore, according to the present analysis, whether or not one adopts an instrumental stance towards the robot's mental states need not affect their perception of robot human likeness. What is claimed here is that perceptions of a robot's psychological human likeness may be affected, among other factors, by the difference between one's folk-ontological stance towards the robot and other human beings. Perception of psychological human likeness is therefore, in the present analysis, a matter of belief, not acceptance. If the focus of HRI research is more on people's everyday 'gut feeling' and spontaneous perception of robots than on their deliberate and philosophically justified judgments about robots' mind, then considerations about what people believe are more relevant to the point than considerations on what they accept.

So far it has been assumed that John's folk-ontological stance towards his fellow human beings is of the psychologically realist variety. But it need not be. The taxonomy formulated before may help one identify various forms of dehumanization, a phenomenon that has been extensively studied (see, for example, [67], [68]). Dehumanization may correspond to believing that human beings do not possess mental states (eliminativism), that they possess non-mental states that may be verbally expressed in mentalistic terms (reductionism), that they possess mental states in the same sense in which Sherlock Holmes believes that he lives in Baker Street (fictionalism). An analysis of the phenomenon of dehumanization exceeds the scope of this article. What is suggested here is that people's perception of robot human likeness do not depend only on their folk-ontological stance towards robots, but also on their folk-ontological stance towards human beings. If John is eliminativist towards robots *and* humans, he will likely perceive the robot as human-like.

To sum up. If the perception of robots' psychological human likeness depends on the relation between people's mental models of human and robot minds, then people's folk-ontological stances may make the difference in their perception of robots' psychological human likeness. And if the latter factor can shape the dynamics of HRI, then the theoretical and empirical study of people's folk-ontological stances towards robots must be relevant to HRI research. This is another reason to believe that research on mental state attribution to robots should make an 'folk-ontological turn'.

---

[12] Admittedly, this analysis somehow presupposes that the perception of a robot's psychological human likeness is an all-or-nothing matter, i.e., that one may perceive the robot as human-like or not. This is an oversimplification, and it is reasonable to require that any plausible conceptual account of psychological human likeness will assume that human likeness can admit of degrees. This request will not be fulfilled in this article. However, the analysis proposed here is in principle compatible with it. People's mental models of robots may differ from their mental models of fellow humans in varying respects and degrees, and this may be reflected in the perception of one robot as more or less similar to a human than another.

## 5. Summary and conclusive remarks

Based on an analysis of the notions of 'attribution' and 'ontological commitment', a repertoire of possible folk-ontological stances towards robots, plus one folk-epistemological stance, have been identified. During the course of this analysis, it has been proposed that to attribute a (mental) state to robot is tantamount to believing that that robot has that mental state, and that, in this perspective, one cannot attribute mental states to robots and at the same time *not* believe that that robot really has mental states. In more general terms, one cannot attribute mental states to robots and fail to take a folk-ontological stance about the robot's mind. Unless one adopts a different conception of 'mental state attribution', it is not clear how the so-called 'attribution question' and 'belief question' can be distinguished from one another. Thus, one way in which this article intends to contribute to HRI research is by offering a conceptual analysis of the notion of 'attribution' which is typically used as primitive in the literature, and by reflecting on the tenability of the distinction between the 'belief' and the 'attribution' questions. Another specific message this article intends to convey is that to attribute a mental state to a robot is not the same thing as to make an instrumental use of mentalistic notions to explain or predict its behaviour. The latter case corresponds to the deliberate acceptance of particular assumptions about the robot, while attribution is a matter of beliefs. It is possible that people deliberately assume that robots have mental states, but this does not mean, *per se*, that they are attributing mental states to them.

Few studies have so far *explicitly* pursued the analysis of what are here called folk-ontological stances of people towards robots. In this context it has been argued that a 'folk-ontological turn' in HRI research could lead to a deeper understanding of people's perception of robots. It could also shed light on people's perception of robots' psychological human likeness, which depends, among other factors, on the difference between the mental model of a robot's mind and the mental model of the human mind. Since folk-ontological stances are part of people's mental models of robots and humans, they are likely to modulate their perception of robots' human likeness. And, if people's perception of robots' human likeness can influence how they react to robots, then their folk-ontological stances towards robots can shape the dynamics of HRI. This article has attempted to offer reasons to believe that studying people's folk-ontological stances towards robots is possible and relevant to HRI research, and that philosophy of science and mind can play a crucial role in this theoretical and experimental undertaking.

On the other way around, this article can also be read as promoting a deeper reflection on the notions of 'attribution', 'ontological commitment', and related foundational concepts in HRI research. Most claims made here rely on a particular analysis of the concept of 'attribution'. The distinction between the belief and the attribution question, and the thesis that one can attribute mental states to robots without making any ontological commitments to the reality of those states, could be defended by employing a different analysis of attribution and ontological commitment. This article, as such, issues a conceptual challenge to the HRI community.

One huge problem that this article does not address is how folk-ontological stances could be investigated. Mental state attributions to robots are currently studied through experimental tools such as questionnaires [23], [27], [28], [29], [69], non-verbal behavioural measures (e.g., detection of anticipatory gaze as in [70]), and neurological techniques (such as fMRI, as in [61]). Still, Thellman and colleagues [3] argue that "A particularly pressing issue is that there is so far very little explicit discussion about what kinds of data constitute evidence of mental state attribution to robots". More importantly for the present purposes, the methods developed so far seem to be inadequate to reveal people's folk-ontological stances towards robots. Fussell and colleagues [30], for example, claim that "A question remains as to whether participants' judgments in the reaction time study reflect beliefs that robots literally possess feelings, attitudes and personality traits or whether they instead are based on metaphoric extension. This question cannot be determined by reaction time data alone, as studies have shown that evaluations of metaphoric statements like 'my surgeon is a butcher' can be as rapid as that of literal statements like 'John is a butcher', especially when the prior context supports the metaphorical interpretation". This problem is not addressed here. However, it is worth stressing that, in the perspective developed here, people's folk-ontological stances consist in their beliefs (or, equivalently, in their attributions) and are integral parts of their mental models of robots. In principle, there is no deep distinction between the problem of determining people's mental models of robots and the problem of determining their folk-ontological stance towards them: it is all a matter of what they believe. Their attributions are underdetermined by their utterances – or reaction times, or answers to questionnaires – and so are their folk-ontological stances. There may still be quite a long way to go towards a full understanding of people's understanding of robots, but from the analysis carried out here it follows that the way is already paved for the 'folk-ontological turn' envisaged here.